\documentclass[conference]{IEEEtran}
\IEEEoverridecommandlockouts
\usepackage{cite}
\usepackage{amsmath,amssymb,amsfonts}
\usepackage[ruled,vlined]{algorithm2e}
\usepackage{multirow}
\usepackage{wrapfig}
\usepackage{booktabs}
\usepackage[para,online,flushleft]{threeparttable}
\usepackage{graphicx}
\usepackage{textcomp}
\usepackage{xcolor}
\usepackage{caption}
\DeclareCaptionLabelFormat{IEEEfig}{Fig.\ \thefigure.}
\DeclareCaptionLabelFormat{IEEETable}{TABLE\ \Roman{table}}
\usepackage{titlesec}

\newtheorem{assumption}{Assumption}
\newtheorem{theorem}{Theorem}
\newtheorem{remark}{Remark}
\titlespacing{\section}{0pt}{1pt}{1pt}
\titlespacing{\subsection}{0pt}{1pt}{1pt}

\def\BibTeX{{\rm B\kern-.05em{\sc i\kern-.025em b}\kern-.08em
    T\kern-.1667em\lower.7ex\hbox{E}\kern-.125emX}}

\makeatletter
\newcommand{\linebreakand}{%
  \end{@IEEEauthorhalign}
  \hfill\mbox{}\par
  \mbox{}\hfill\begin{@IEEEauthorhalign}
}
\makeatother

\begin{document}
\thispagestyle{empty}
\pagestyle{empty}

\title{
\LARGE \textbf{\texttt{FUSE}: First-Order and Second-Order Unified \\SynthEsis in Stochastic Optimization}}
\author{\IEEEauthorblockN{Zhanhong Jiang$^1$, Md Zahid Hasan$^1$, Aditya Balu$^1$, Joshua R. Waite$^1$, Genyi Huang$^2$, Soumik Sarkar$^1$ \\
\textit{$^1$Iowa State University, $^2$Oracle}\\
$\{\text{zhjiang, zahid, baditya, jrwaite, soumiks}\}$@iastate.edu, genyi.huang@oracle.com
}
}

\maketitle

\begin{abstract}
Stochastic optimization methods have actively been playing a critical role in modern machine learning algorithms to deliver decent performance. While numerous works have proposed and developed diverse approaches, first-order and second-order methods are in entirely different situations. The former is significantly pivotal and dominating in emerging deep learning but only leads convergence to a stationary point. However, second-order methods are less popular due to their computational intensity in large-dimensional problems.
This paper presents a novel method that leverages both the first-order and second-order methods in a unified algorithmic framework, termed \texttt{FUSE}, from which a practical version (PV) is derived accordingly. \texttt{FUSE-PV} stands as a simple yet efficient optimization method involving a switch-over between first and second orders. Additionally, we develop different criteria that determine when to switch. \texttt{FUSE-PV} has provably shown a smaller computational complexity than SGD and Adam. To validate our proposed scheme, we present an ablation study on several simple test functions and show a comparison with baselines for benchmark datasets.
\end{abstract}


\section{Introduction}
Modern machine learning has dramatically benefited from stochastic optimization algorithms that provide highly effective iterative methods to attain optimal solutions. 
The emerging deep learning particularly motivates the broad development and applications of these methods. Among them, first-order stochastic optimization approaches have gained considerable attention. Stochastic gradient descent (\texttt{SGD})~\cite{bottou2012stochastic} has provably and empirically shown its pronounced efficiency and efficacy for various problems. Albeit adopted widely, \texttt{SGD} suffers from slow convergence behavior that inevitably increases the computational complexity. To mitigate this issue, the technique of adaptive moment has already been utilized to yield another class of first-order methods, such as \texttt{AdaGrad}~\cite{defossez2020convergence}, \texttt{RMSProp}~\cite{zou2019sufficient} and \texttt{Adam}~\cite{defossez2020convergence,zou2019sufficient}. Such adaptive gradient methods are primarily driven by applying the first and/or second moments of gradients into the gradient descent step, leading to adaptive step sizes. Nevertheless, this causes the issue of \textit{poor generalization}, regardless of the empirically faster convergence compared to \texttt{SGD}. 
More recently, advanced optimizers, e.g., \texttt{AdaBound}~\cite{savarese2019convergence} and \texttt{RAdam}~\cite{liu2019variance} were also developed to enhance the training performance and testing accuracy, but their applicability to different types of problems may not necessarily be on par with that of \texttt{Adam}, which thus remains the most popular one. However, all first-order methods, in theory, can only guarantee the convergence to the \textit{first-order stationary point} (FOSP) under the first-order necessary condition~\cite{boyd2004convex}. 

Another line of work of interest that has not thoroughly been investigated is second-order stochastic optimization, which leverages the second-order information of the objective loss. Compared to first-order algorithms, second-order algorithms calculate or approximate the Hessian matrix and/or its inverse, resulting in a large computational overhead. Moreover, empirically, some popular second-order methods, such as Limited-memory Broyden–Fletcher–Goldfarb–Shanno (\texttt{L-BFGS})~\cite{zhu1997algorithm} and Conjugate Gradient (\texttt{CG})~\cite{polyak1969conjugate}, still underperform first-order methods particularly in solving deep learning-based optimization problems. 
However, second-order approaches retain the convergence to the \textit{local minimum}, possibly with a much faster convergence rate given the proper initialization. Recently, some works~\cite{xu2020second,yao2020adahessian,anil2020scalable,anil2020second, ma2020apollo,jin2021exploiting,di2023lsos,arjevani2020second} have attempted to develop effective second-order schemes for training overparameterized models in deep learning. A more recent work~\cite{niu2023ml} adopted \texttt{SGD} as a warmup for some iterations before applying a variant of \texttt{L-BFGS}, but this work was under the distributed setting.

To alleviate the generalization error while retaining the fast convergence in training over-parameterized models, a previous work~\cite{keskar2020improving} combined \texttt{Adam} and \texttt{SGD} together by defining a switchover point that determined when to switch from \texttt{Adam} to \texttt{SGD}. 
More recent works~\cite{landro2020mixing,wang2020adasgd} have also reported similar ideas to mix these two optimizers. However, the guarantee of local minimum is still not necessarily true due to the property of FOSP. In this work, we take another way of combination between first-order and second-order methods that has not been explored and developed. Specifically, this paper presents the following contributions.
a) We develop a novel algorithm framework, termed \texttt{FUSE} (First-order and second-order Unified SynthEsis), for solving the stochastic optimization problems in machine learning, through combining the first-order adaptive gradient method (e.g., \texttt{Adam}) with the second-order method (e.g., \texttt{L-BFGS}). 
b) We derive a practical version, termed \texttt{FUSE-PV}, which involves a ``switchover" point between first-order and second-order methods. We define three metrics, iteration (epoch)-based, gradient-based and loss-based for the implementation. \texttt{FUSE-PV} provably shows the improvement of computational complexity compared to \texttt{SGD} and \texttt{Adam}. 
    Please also see Table~\ref{table:comparison} for comparison.
c) To validate the proposed algorithm, we utilize simple test functions and popular benchmark datasets with multiple models. The empirical results show the superiority of \texttt{FUSE-PV} over other baseline optimizers. 
\begin{table}[htp]
\caption{Comparison among different methods.}
\begin{center}
\begin{threeparttable}
\begin{tabular}{c c c c}
    \toprule
    \textbf{Method} & \textbf{function} & \textbf{Order} & \textbf{Complexity}\\ \midrule
      \texttt{SGD}   & NC& First                   &  $\mathcal{O}(\frac{1}{\epsilon^2})$   \\
      \texttt{Adam}   & NC& First                     &  $\mathcal{O}(\frac{1}{\epsilon^2})$           \\ \hline
      \multirow{2}{4em}{\texttt{L-BFGS}} & SC& Second & $\mathcal{O}(\textnormal{log}\frac{1}{\epsilon})$ \\&NC&Second&$\mathcal{O}(\frac{1}{\epsilon^2})$\\\hline
      \multirow{2}{4em}{\texttt{FUSE-PV}} & SC & F+S & $\mathcal{O}(\textnormal{max}\{\frac{1}{\zeta},\textnormal{log}(\frac{1}{\epsilon})\})$\\&NC&F+S&$\mathcal{O}(\frac{1}{\epsilon^2})$\\
      \bottomrule
      
\end{tabular}
\begin{tablenotes}
NC: non-convex, SC: strongly convex; F+S: First + Second; $\zeta > \epsilon >0$. Please note that due to the stochastic gradient, the \texttt{L-BFGS} and \texttt{FUSE-PV} are in a stochastic version. 
\end{tablenotes}
\end{threeparttable}
\end{center}
\label{table:comparison}
\end{table}

\section{Preliminaries and Proposed Algorithm}
We introduce the preliminary background knowledge for first-order and second-order methods, which will characterize the proposed algorithmic frameworks presented later. In this context, we select \texttt{Adam} and \texttt{L-BFGS} for representatives while deferring other algorithms to future work. Recalling the empirical risk minimization, we have
\begin{equation}\label{erm}
    \textnormal{min}_{\mathbf{x}\in\mathbb{R}^d}f(\mathbf{x}):=\frac{1}{n}\sum_{i\in\mathcal{D}}f^i(\mathbf{x}),
\end{equation}
where $\mathcal{D}$ is the dataset, $n$ is the size of $\mathcal{D}$, $f:\mathbb{R}^d\to\mathbb{R}$ is the loss function and $f^i$ is the function value corresponding to a sample $i$. Depending on concrete problems, $f$ possesses different properties, such as convexity or non-convexity, which has an impact on the convergence rate of the designed algorithm. It has been well-known that most first-order algorithms attain the $\epsilon$-optimality within $\mathcal{O}(\epsilon^{-2})$ iterations based on $\mathbb{E}[\|\nabla f(\mathbf{x}) \|^2]\leq \epsilon$. $\|\cdot\|$ signifies the Euclidean norm.
\subsection{\texttt{Adam} and \texttt{L-BFGS}}
Despite other variants, \texttt{Adam} can be regarded as the most popular stochastic optimizer. By leveraging the exponential moving average to the stochastic gradient and its square, \texttt{Adam} resorts to the first-order ($\mathbf{m}_k$) and second-order ($\mathbf{v}_k$) moments for adjusting step size during optimization. Specifically, the update law of \texttt{Adam} can be framed as follows:
\begin{subequations}\label{adam}
\begin{align}
    \mathbf{m}_{k+1} &= \beta_1 \mathbf{m}_k + (1-\beta_1)\mathbf{g}(\mathbf{x}_k)\\
    \mathbf{v}_{k+1} &= \beta_2 \mathbf{v}_k + (1-\beta_2)\mathbf{g}(\mathbf{x}_k)\odot\mathbf{g}(\mathbf{x}_k)\\
    \mathbf{x}_{k+1} &= \mathbf{x}_k -  \alpha\frac{\sqrt{1-\beta_2^k}}{1-\beta_1^k}\mathbf{m}_{k+1}\oslash\sqrt{\mathbf{v}_{k+1}+a},
\end{align}
\end{subequations}
where $\mathbf{g}(\mathbf{x})$ is the stochastic gradient (we will drop stochastic for simplicity) and it is assumed to be the unbiased estimate of $\nabla f(\mathbf{x})$, i.e., $\nabla f(\mathbf{x})=\mathbb{E}[\mathbf{g}(\mathbf{x})]$, $a$ is a user-defined small positive constant and $k$ is the time step. $\odot$ is the Hadamard product and $\oslash$ is the element-wise division. 
The term in this context $\frac{\sqrt{1-\beta_2^k}}{1-\beta_1^k}$ serves for bias correction due to the exponential moving average in the algorithm. In terms of theoretical and empirical findings, \texttt{Adam} enables a faster convergence, in particular at the beginning of training. However, after a sufficient number of iterations, it converges to a FOSP, which can be implied by the saturation from the training loss. 

According to the previous discussion, \texttt{SGD} requires acceleration for effectively solving optimization problems, and \texttt{Adam} has shown one valid technique. Alternatively, the information from the Hessian matrix can also be used to accelerate the convergence and improve the performance. \texttt{L-BFGS} is in between gradient descent and Newton's method, as it approximates the Hessian without direct calculation, leading to faster convergence and the reduction of high computational cost. We first review the updated laws for \texttt{L-BFGS}. Suppose that the starting point is $\mathbf{x}_0$. \texttt{L-BFGS} involves a two-loop recursion that requires gradient, parameter difference and gradient difference as the input. Denote by $\mathbf{s}_k$ and $\mathbf{y}_k$ the parameter and gradient difference respectively at time step $k$ such that $\mathbf{s}_k=\mathbf{x}_{k+1}-\mathbf{x}_k$ and $\mathbf{y}_k=\mathbf{g}_{k+1}-\mathbf{g}_k$. 
The key of \texttt{L-BFGS} is to calculate a new direction $\mathbf{p}_k$ to update the parameter $\mathbf{x}_{k+1} = \mathbf{x}_k+\alpha_k\mathbf{p}_k$ by using current gradient $\mathbf{g}_k$, the historical $m$ states of $\mathbf{s}$ and $\mathbf{y}$. Specifically, the direction $\mathbf{p_k}$ is obtained in Algorithm~\ref{alg:lbfgs_2}. 

\begin{algorithm}
\caption{\texttt{L-BFGS} Algorithm Outline}\label{alg:lbfgs_1}
\SetAlgoLined
\KwData {starting point $\mathbf{x}_0$, integer history size $m>0$, $k=1$}
\KwResult{$\mathbf{x}^*$}
$k=0$\;
\While{no converge}{
Calculate gradient $\mathbf{g}_k$ at position $\mathbf{x}_k$\;
Compute direction $\mathbf{p}_k$ using Algorithm~\ref{alg:lbfgs_2}\;
Compute $\mathbf{x}_{k+1} = \mathbf{x}_k+\alpha_k\mathbf{p}_k$ where $\alpha_k$ is chosen to satisfy Wolfe conditions\;
\If{$k>m$}{Discard vector pair $\mathbf{s}_{k-m}$, $\mathbf{y}_{k-m}$ from memory storage}
$\mathbf{s}_k=\mathbf{x}_{k+1}-\mathbf{x}_k$,
$\mathbf{y}_k=\mathbf{g}_{k+1}-\mathbf{g}_k$\;
$k=k+1$\;
}
\end{algorithm}

\begin{algorithm}
\caption{\texttt{L-BFGS} two-loop recursion }\label{alg:lbfgs_2}
\SetAlgoLined
\KwData {$\mathbf{g}_k$, $\mathbf{s}_i$, $\mathbf{y}_i$, $i=k-m, ..., k-1$}
\KwResult{new direction $\mathbf{p}_k$}
$\mathbf{p}=-\mathbf{g}_k$\;
\For{$i=k-1:k-m$}{$\alpha_i = \frac{\mathbf{s}_i\mathbf{p}}{\mathbf{s}_i\mathbf{y}_i}$, $\mathbf{p=p-\alpha_i\mathbf{y}_i}$}
$\mathbf{p}=(\frac{\mathbf{s}_{k-1}\mathbf{y}_{k-1}}{\mathbf{y}_{k-1}\mathbf{y}_{k-1}})\mathbf{p}$\;
\For{$i=k-1:k-m$}{$\beta = \frac{\mathbf{y}_i\mathbf{p}}{\mathbf{s}_i\mathbf{y}_i}$, $\mathbf{p}=\mathbf{p}+(\alpha_i-\beta)\cdot\mathbf{s}_i$}
\end{algorithm}

In Algorithm~\ref{alg:lbfgs_1} the gradient $\mathbf{g}_{k}$ is calculated in a mini-batch way such that it is equal to $\frac{1}{|\mathcal{B}|}\sum_{i\in\mathcal{B}}\nabla f^i(\mathbf{x}_{k})$, where $\mathcal{B}$ is a mini-batch of $\mathcal{D}$. \texttt{L-BFGS} has a significant advantage when solving linear/quadratic minimization problems as the convergence is in finite steps. However, modern machine learning problems are naturally highly nonlinear and non-convex, particularly when they are overparameterized neural networks. Due to the two-loop recursion, it could probably take a long time for \texttt{L-BFGS} to converge to a decent optimum. For instance, if $K$ iterations must be run before satisfying the convergence criterion, then the total run time cost is $\mathcal{O}(Kmd)$. For a large deep neural network model, this could be extremely time-consuming. Our work presents an approach to reduce such computational overhead.
\subsection{First-order and second-order Unified SynthEsis (\texttt{FUSE})}
The proposed \texttt{FUSE} is shown in Algorithm~\ref{alg:FUSE-1}. One observation from Algorithm~\ref{alg:FUSE-1} is that \texttt{FUSE} utilizes a parameter to control the update for $\mathbf{x}$, which can be treated as a ``soft" update by using $\mathbf{x}^A_{k+1}$ and $\mathbf{x}^L_{k+1}$ that are obtained from \texttt{Adam} and \texttt{L-BFGS} respectively. $\theta$ provides a convenient way to put importance to different algorithms at the current epoch. A good practice is in the early phase of optimization, $\theta$ can be close to 1 and then gradually reduce to 0. When approaching the optimal solution, \texttt{L-BFGS} reduces the number of iterations. Another observation is that \texttt{FUSE} has more iteration costs compared to either \texttt{Adam} or \texttt{L-BFGS} only. During each iteration, it needs to calculate the parameter updates using both first-order and second-order methods. When $d$ is large, \texttt{FUSE} could become computationally intractable. Thus, in this context, we derive a practical version of \texttt{FUSE}, i.e., \texttt{FUSE-PV}, which introduces a \textit{switchover} condition to decide between first-order and second-order algorithms. Equivalently, in \texttt{FUSE-PV}, two phases are established based on the switchover point and $\theta$ is selected from $\{0,1\}$. The algorithmic framework is in Algorithm~\ref{alg:FUSE-2}. 
The switchover point can be determined by any proper condition. We defer the explicit expression to the later section. While \texttt{FUSE-PV} simply allows \texttt{Adam} and \texttt{L-BFGS} to take control of different stages of the optimization, this procedure has led to a decent empirical performance. During optimization, \texttt{Adam} triggers a fast search for the optimal solution at the early stage, in which both \texttt{SGD} and second-order algorithms can perform poorly. While in the later stage, close to the optimal solution, second-order algorithms are able to take significantly fewer iterations to reach it compared to first-order algorithms, given its additional curvature information. This is equivalent to a scenario in which the initialization point was selected in a proper way such that the convergence is pretty fast. 
We now introduce the switchover condition in what follows. 
\begin{algorithm}
\caption{\texttt{FUSE}: \texttt{Adam} + \texttt{L-BFGS}}\label{alg:FUSE-1}
\SetAlgoLined
\KwData {$\beta_1,\beta_2\in(0,1], \mathbf{x}_0, \alpha, \mathbf{m}_0, \mathbf{v}_0, K, \epsilon, \theta\in [0,1]$}
\KwResult{$\mathbf{x}_K$}
$k=0$\;
\While{$k<K$}{
Update $\mathbf{x}^A_{k+1}$ using Eq.~\ref{adam}\;
Update $\mathbf{x}^L_{k+1}$ using Algorithm~\ref{alg:lbfgs_1}\;
$\mathbf{x}_{k+1} = \theta\mathbf{x}^A_{k+1} + (1-\theta)\mathbf{x}^L_{k+1}$\;
$k=k+1$\;
}
\end{algorithm}

\begin{algorithm}
\caption{\texttt{FUSE-PV}}\label{alg:FUSE-2}
\SetAlgoLined
\KwData {$\beta_1,\beta_2\in(0,1], \mathbf{x}_0, \alpha, \mathbf{m}_0, \mathbf{v}_0, K, \epsilon$}
\KwResult{$\mathbf{x}_K$}
$k=0$\;
\While{$k<K$}{
\eIf{switchover condition is satisfied}
    {Update $\mathbf{x}_{k+1}$ using Algorithm~\ref{alg:lbfgs_1}}{
    Update $\mathbf{x}_{k+1}$ using Eq.~\ref{adam}}
$k=k+1$\;
}
\end{algorithm}

\subsection{Metrics for Switchover}
\noindent\textbf{Simple iteration (epoch).}
A simple method that can be immediately used as a switchover condition is to define how many iterations for first-order and second-order methods to run, respectively. Such a method is easy to implement and has more practical efficiency. Furthermore, it doesn't add any additional computational cost when running the algorithm. However, correctly determining the accurate number of iterations could be a quite challenging problem, which may require some trials. Though one can regard it as a hyperparameter and use grid search or hyperparameter optimization methods to determine, this is out of our study scope.
\noindent\textbf{Gradient norm.}
We introduce another switchover condition that is related to the gradient norm. In an ideal optimization process, the gradient will decay to 0 along with iterations. However, due to stochasticity and the typical requirement of non-asymptotic convergence, the gradient only converges to a sufficiently small value within an acceptable accuracy. According to this, we have
\begin{equation}\label{eq_3}
    \frac{1}{K}\sum_{k=1}^K\mathbb{E}[\|\mathbf{g}(\mathbf{x}_k)\|^2]\leq\zeta,
\end{equation}
where $\zeta$ is a user-defined positive constant. This condition guarantees that when activating the second-order algorithm, $\mathbf{x}_k$ has been properly initialized with the first-order algorithm, even close to $\mathbf{x}^*$. It has been well-known that first-order algorithms, such as \texttt{SGD}, can suffer from slow convergence when $\mathbf{x}_k\to\mathbf{x}^*$, while second-order algorithms have superior performance in this region. Thus, the gradient norm condition with a properly defined $\zeta$ facilitates the convergence to the optimal solution. Though Eq.~\ref{eq_3} can only guarantee local optimality, it has been adopted quite often in non-convex optimization. When applying the gradient norm condition, we are not aware of how many times the value of $\zeta$ is the final accuracy (say $\epsilon$). While $\zeta$ should be defined as larger than $\epsilon$, their relationship is one future work to be investigated.
\noindent\textbf{Objective loss.}
The final switchover condition in this work we want to introduce is to leverage the loss. When evaluating an optimization algorithm, one can observe the loss directly. Generically, we would like the designed algorithm to keep descending the loss when searching for a minimum. Hence, the difference between two consecutive losses can be used to determine when to switch over from the first-order to the second-order methods, i.e.,
\begin{equation}
    \mathbb{E}[|f(\mathbf{x}_k) - f(\mathbf{x}_{k-1})|]\leq \sigma,
\end{equation}
where $\sigma$ is also a user-defined small positive constant.
However, this condition has relied on an implicit fact that the gradient is bounded in optimization. We also assume $f^*=f(\mathbf{x}^*)>-\infty$ throughout the rest of the analysis, where $\mathbf{x}^*:=\textnormal{argmin}_{\mathbf{x}\in\mathbb{R}^d}f(\mathbf{x})$. Other conditions such as $\mathbb{E}[\|\mathbf{x}_k-\mathbf{x}^*\|]\leq\sigma$ or $\mathbb{E}[|f(\mathbf{x}_k)-f(\mathbf{x}^*)|]\leq\sigma$ could also be used as the switchover condition, but typically $\mathbf{x}^*$ is unknown a priori.

\section{Theoretical Analysis}
In this section, we present theoretical analysis for the proposed \texttt{FUSE-PV} algorithm and leave the more general analysis of \texttt{FUSE} in our future work. Due to the space limit, we only present the proof sketch for the non-convex case, skipping the strongly convex one.
We first present some assumptions to characterize the main results. 
\begin{assumption}\label{assump_1}
    (a) $f^i$ is smooth with constant $L>0$ and twice continuously differentiable, for all $i$; (b)
    There exist constants $\eta>0$ and $\gamma\geq0$ such that $\mathbb{E}_{i\sim\mathcal{D}}[\|\nabla f^i(\mathbf{x})\|^2]\leq \gamma^2+\eta\|\nabla f(\mathbf{x})\|$.
\end{assumption}
Assumption~\ref{assump_1} (b) implies the bounded variance that has been generic in previous works~\cite{shi2020rmsprop,bottou2018optimization,berahas2016multi} and reduces to strong growth condition if $\gamma=0$~\cite{vaswani2019fast}. Another popular assumption is bounded gradient when showing the proof for adaptive gradient algorithms. The authors~\cite{shi2020rmsprop} have presented proof techniques without this stronger assumption, which can be violated easily even with a simple quadratic loss. We first analyze the complexity of the strongly convex functions.
\begin{theorem}(Informal)\label{theorem_1}
    Suppose that Assumption~\ref{assump_1} holds and that $f^i$ is strongly convex with constant $\mu>0$. There exists constant $\zeta>0$ and $\epsilon>0$ such that $\zeta>\epsilon$. Then \texttt{FUSE-PV} incurs the complexity with the order of $\mathcal{O}(\textnormal{max}\{\frac{1}{\zeta},\textnormal{log}(\frac{1}{\epsilon})\})$.
\end{theorem}
\begin{remark}\label{remark1}
In Theorem~\ref{theorem_1}, $\mathcal{O}(\frac{1}{\zeta})$ is due to \texttt{Adam}, which matches the result in~\cite{wang2019sadam}, and this still holds if other first-order algorithms are used in this context such as \texttt{SGD}. The faster component $\mathcal{O}(\textnormal{log}(\frac{1}{\epsilon}))$ is attributed to \texttt{L-BFGS}. Such an order is obtained by using a \textit{constant} step size, which leads to convergence to the neighborhood of the optimal solution. One can instead utilize a diminishing step size to achieve $\mathbf{x}^*$ but with the cost of a larger computational overhead. Without stochasticity, \texttt{FUSE-PV} can even attain a better convergence with only $\mathcal{O}(\textnormal{max}\{\textnormal{log}(\frac{1}{\zeta}),\textnormal{log}\textnormal{log}(\frac{1}{\epsilon})\})$, which will be validated by empirical evidence. We next give another main result for the non-convex case.
\end{remark}
\begin{theorem}(Informal)\label{theorem_2}
    Suppose that Assumption~\ref{assump_1} holds. There exists constant $\epsilon>0$ such that \texttt{FUSE-PV} incurs the complexity with the order of $\mathcal{O}(\frac{1}{\epsilon^2})$.
\end{theorem}
\textit{Proof sketch:} Denote by $K_1$ the number of epochs for \texttt{Adam} and $K_2$ the number of epochs for \texttt{L-BFGS}. To quantify the complexity for \texttt{Adam}, we adopt the metric $\frac{1}{K_1}\sum_{k=0}^{K_1-1}\mathbb{E}[\|\nabla f(\mathbf{x}_{k})\|^2]\leq \epsilon$. We first use the smoothness condition to obtain the descent lemma for $f(\mathbf{x})$. Subsequently, following the proof techniques from~\cite{defossez2020convergence}, we can obtain $\frac{1}{K_1}\sum_{k=0}^{K_1-1}\mathbb{E}[\|\nabla f(\mathbf{x}_{k})\|^2]\leq\mathcal{O}(\frac{1}{\sqrt{K_1}})$. With the initialization $\mathbf{x}_{K_1}$ for \texttt{L-BFGS}, we have the initialization error as $f(\mathbf{x}_{K_1})-f^*$. We still establish the similar descent lemma by using Assumption~\ref{assump_1}. With the analogous proof techniques from~\cite{ma2020apollo}, we have $\frac{1}{K_2}\sum_{k=0}^{K_2-1}\mathbb{E}[\|\nabla f(\mathbf{x}_{k})\|^2]\leq\mathcal{O}(\frac{\textnormal{log}K_2}{\sqrt{K_2}})$. Hence, the combined complexity is roughly $\mathcal{O}(\frac{1}{\epsilon^2})$.
\begin{remark}\label{remark2}
Theorem~\ref{theorem_2} suggests that generically, if the objective loss is non-convex, then the complexity is similar to that attained by most first-order algorithms. However, to make the most advantage of \texttt{L-BFGS}, a larger batch size can be employed in training to boost the performance, which has also been stressed in~\cite{berahas2016multi}. Likewise, a constant step size is adopted to get $\mathcal{O}(\frac{1}{\epsilon^2})$. A diminishing step size leads to the convergence to the optimal solution, but with the compromise of convergence rate, in an asymptotic manner. 
\end{remark}

\section{Empirical Results}
To validate the proposed approach, we use both low-dimensional non-convex functions and real datasets with multiple models (logistic regression, multi-layer perceptron (MLP), convolutional neural network (CNN) (2 layers), and Densely Connected Convolutional Networks (DenseNet) (5 layers)) to show the performance. To compare the proposed optimizer, we use \texttt{SGD}, \texttt{Adam+SGD} and \texttt{Adam} as baselines. For \texttt{SGD} and \texttt{Adam}, their step sizes are respectively 0.01 and 0.001. Though larger models typically result in better performance, we are concerned more about the performance w.r.t. various optimizers in this work. The datasets we employ consists MNIST, FashionMNIST, KMNIST~\cite{xiao2017fashion}, USPS, and CIFAR 10~\cite{kasun2016dimension}. To keep the indication simple, we drop PV from \texttt{FUSE-PV} in empirical results.

\begin{figure}[ht]
    \centering
    \captionsetup{skip=-1pt}
    \includegraphics[width=8cm]{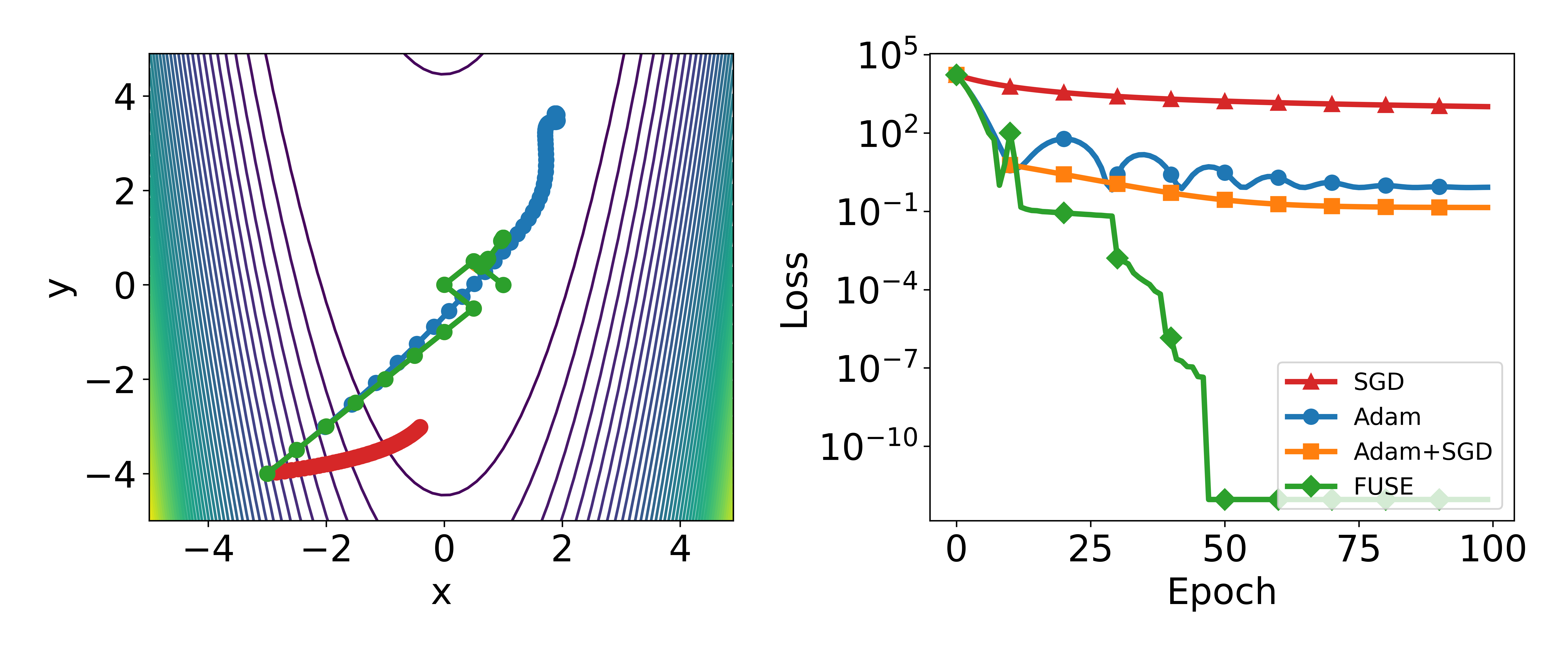}
    \caption{Optimizer performance for 2D Rosenbrock function.}
    \label{fig:rosenbrock}
\end{figure}
\begin{figure}[ht]
    \centering
    \captionsetup{skip=-1pt}
    \includegraphics[width=8cm]{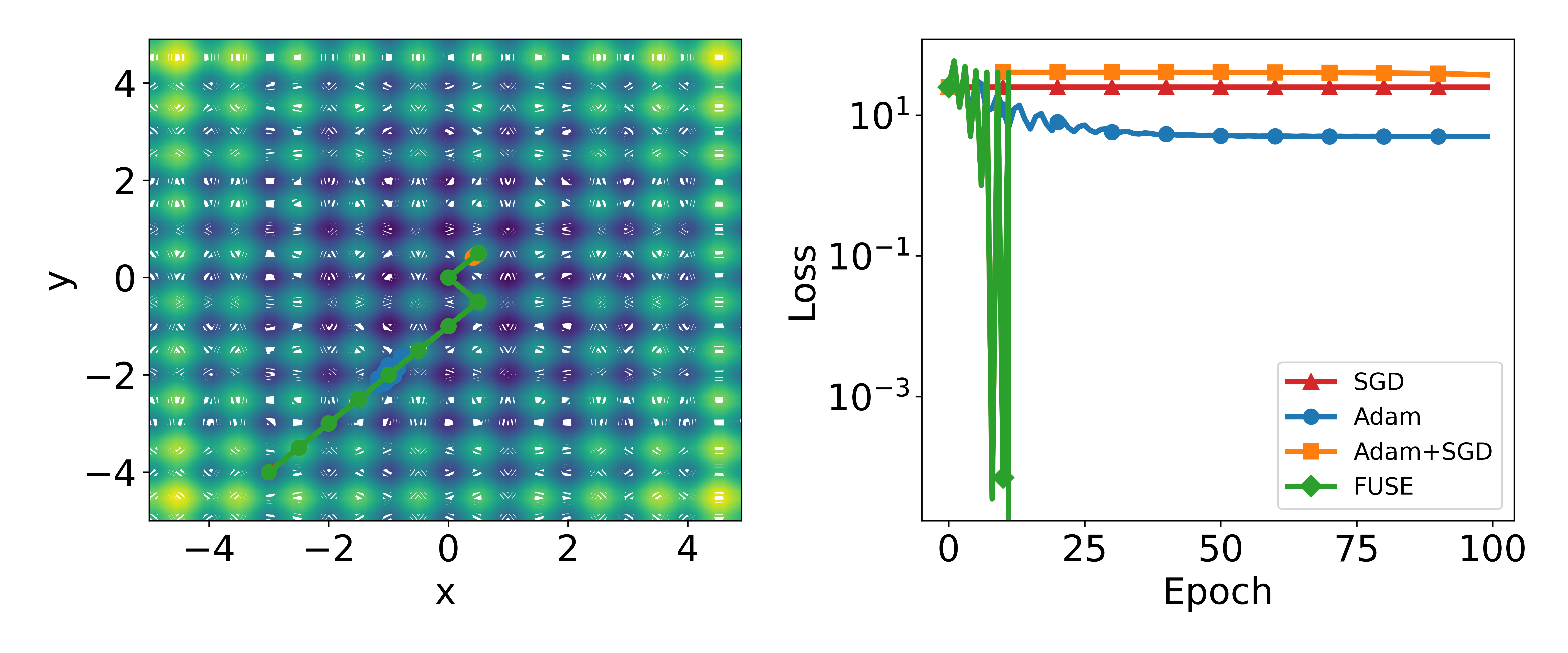}
    \caption{Optimizer performance for 2D Rastrigin function.}
    \label{fig:rastrigin}
\end{figure}
\begin{figure}[ht]
    \centering
    \captionsetup{skip=-1pt}
    \includegraphics[width=8cm]{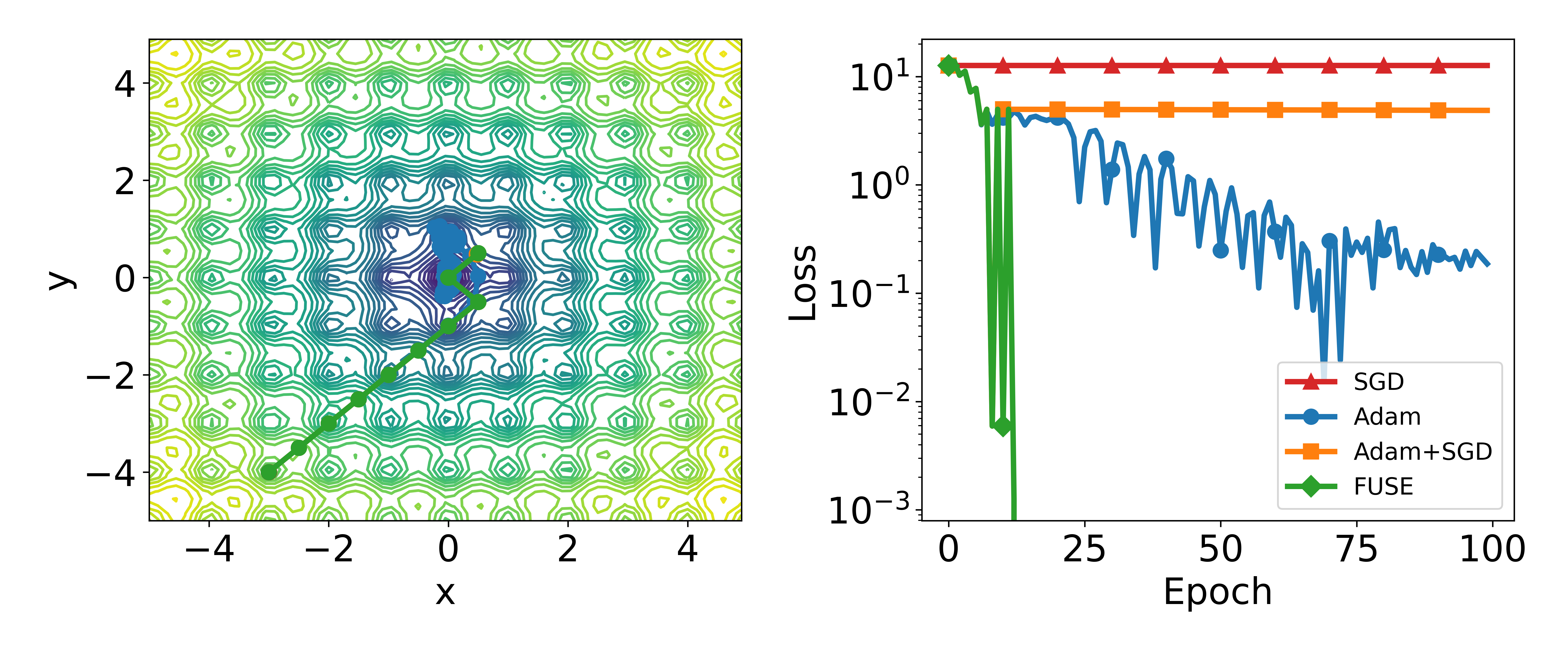}
    \caption{Optimizer performance for 2D Ackley function.}
    \label{fig:ackley}
\end{figure}
\begin{figure}[ht]
    \centering
    \captionsetup{skip=-1pt}
    \includegraphics[width=8cm]{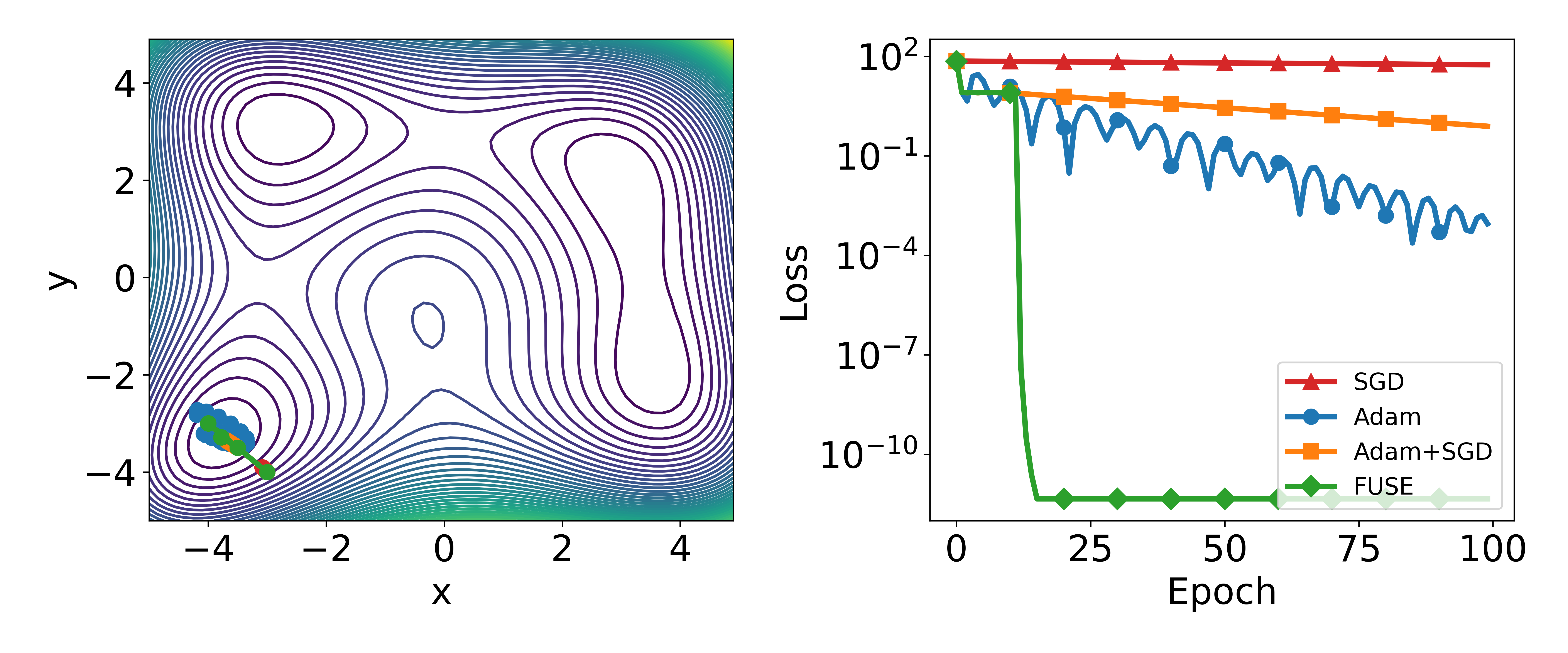}
    \caption{Optimizer performance for 2D Himmelblau function.}
    \label{fig:himm}
\end{figure}

\noindent\textbf{Simple non-convex test function.} 
\begin{table*}[ht]
\captionsetup{skip=-5pt} 
\caption{Performance ($\text{mean}_{\text{std}}$) of different models.}
\begin{center}
\normalsize
\begin{threeparttable}
\begin{tabular}{*{5}{c}}
    \toprule
    \textbf{Dataset/Models} & \textbf{\texttt{SGD}} & \textbf{\texttt{Adam} } & \textbf{\texttt{Adam+SGD}}&
    \textbf{\texttt{FUSE}}\\ 
    \midrule
    MNIST/Logistic Reg.&$0.786_{\sim0.0}$&$0.905_{0.0008}$&$0.894_{\sim0.0}$&$\mathbf{0.908_{0.0042}}$\\
KMNIST/MLP&$0.656_{0.0014}$&$0.788_{0.0064}$&$0.787_{0.0016}$&$\mathbf{0.795_{0.0016}}$\\
CIFAR/DenseNet&$0.150_{0.0024}$&$0.343_{0.0050}$&$0.295_{0.0053}$&$\mathbf{0.366_{0.0100}}$\\
USPS/CNN&$0.229_{\sim0.0}$&$0.899_{0.0056}$&$0.822_{0.0017}$&$\mathbf{0.912_{0.0091}}$\\
    \bottomrule
\end{tabular}
\end{threeparttable}
\end{center}
\label{table:model_results}
\vspace{-17pt}
\end{table*}
Fig.~\ref{fig:rosenbrock} to Fig.~\ref{fig:himm} show different loss landscapes and curves by using optimizers to optimize popular low-dimensional non-convex functions, including 2D Rosenbrock, Rastrigin, Ackley, and Himmelblau functions~\cite{shehab2017survey}.
To have a fair comparison, we set the same initialization for all optimizers in different scenarios. All plots depict the superiority of the proposed \texttt{FUSE} over baselines. Particularly, the switchover from \texttt{Adam} to \texttt{L-BFGS} improves the performance compared to those only using first-order methods, which could get stuck with poor local minimum or have no progress, as observed from Fig.~\ref{fig:rastrigin} and Fig.~\ref{fig:himm}. On the contrary, the currently most popular optimizer \texttt{Adam} saturates after some epochs when solving simple non-convex functions with larger variance. This evidently supports that \texttt{Adam} suffers from a bad stationary point even in low-dimensional problems. From all loss curves plots, \texttt{FUSE} has shown the superlinear convergence without stochastic gradients, validating the theoretical claim in Remark~\ref{remark1}.

\begin{figure}[ht]
    \centering
    \captionsetup{skip=-1pt}
    \includegraphics[width=4cm]{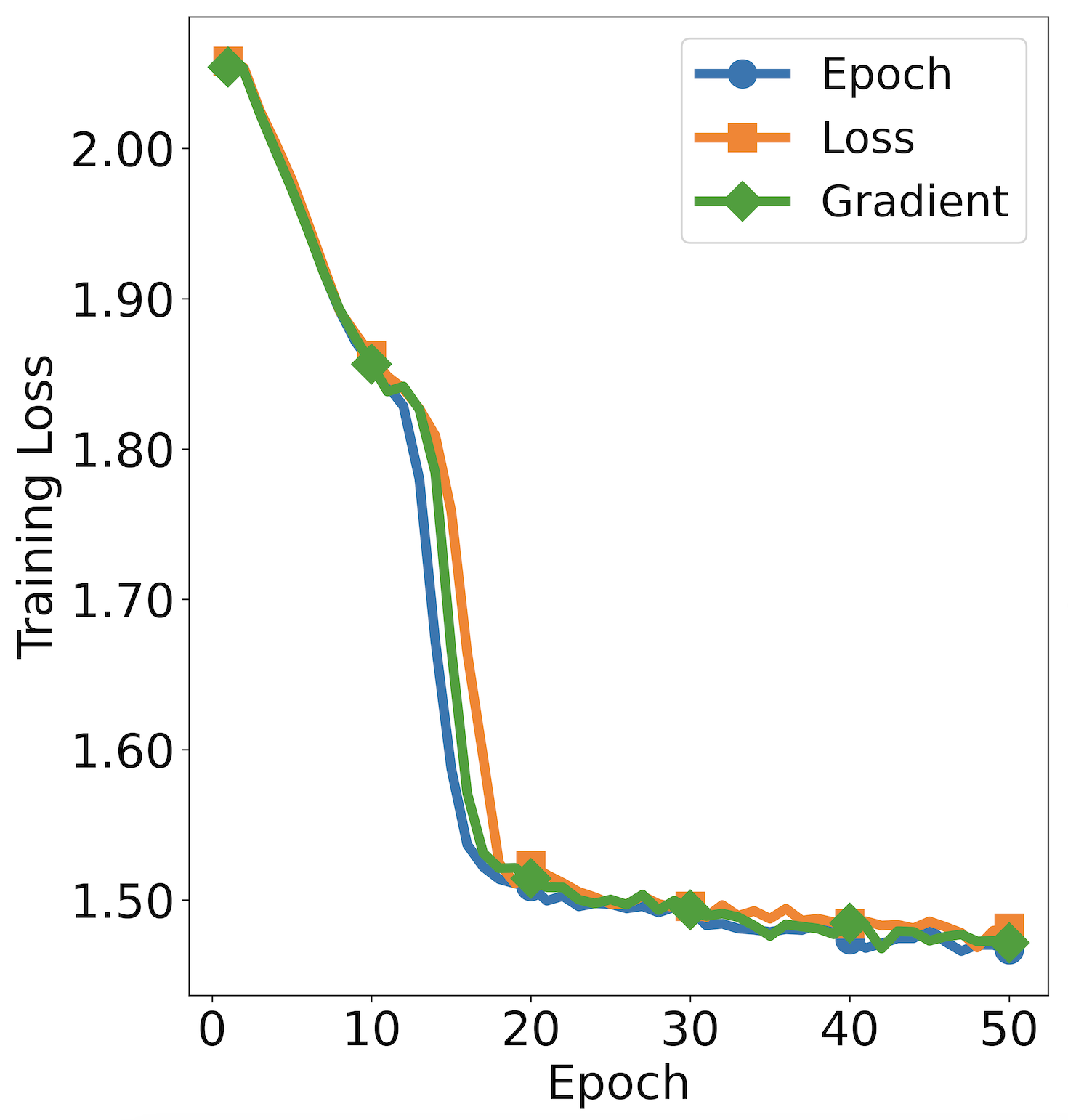}
    \caption{Training loss for different criteria (DenseNet on CIFAR-10)}
    \label{fig:criterion}
\end{figure}

\begin{figure}[hb]
    \centering
    \captionsetup{skip=-1pt}
    \includegraphics[height=4cm, width=7cm]{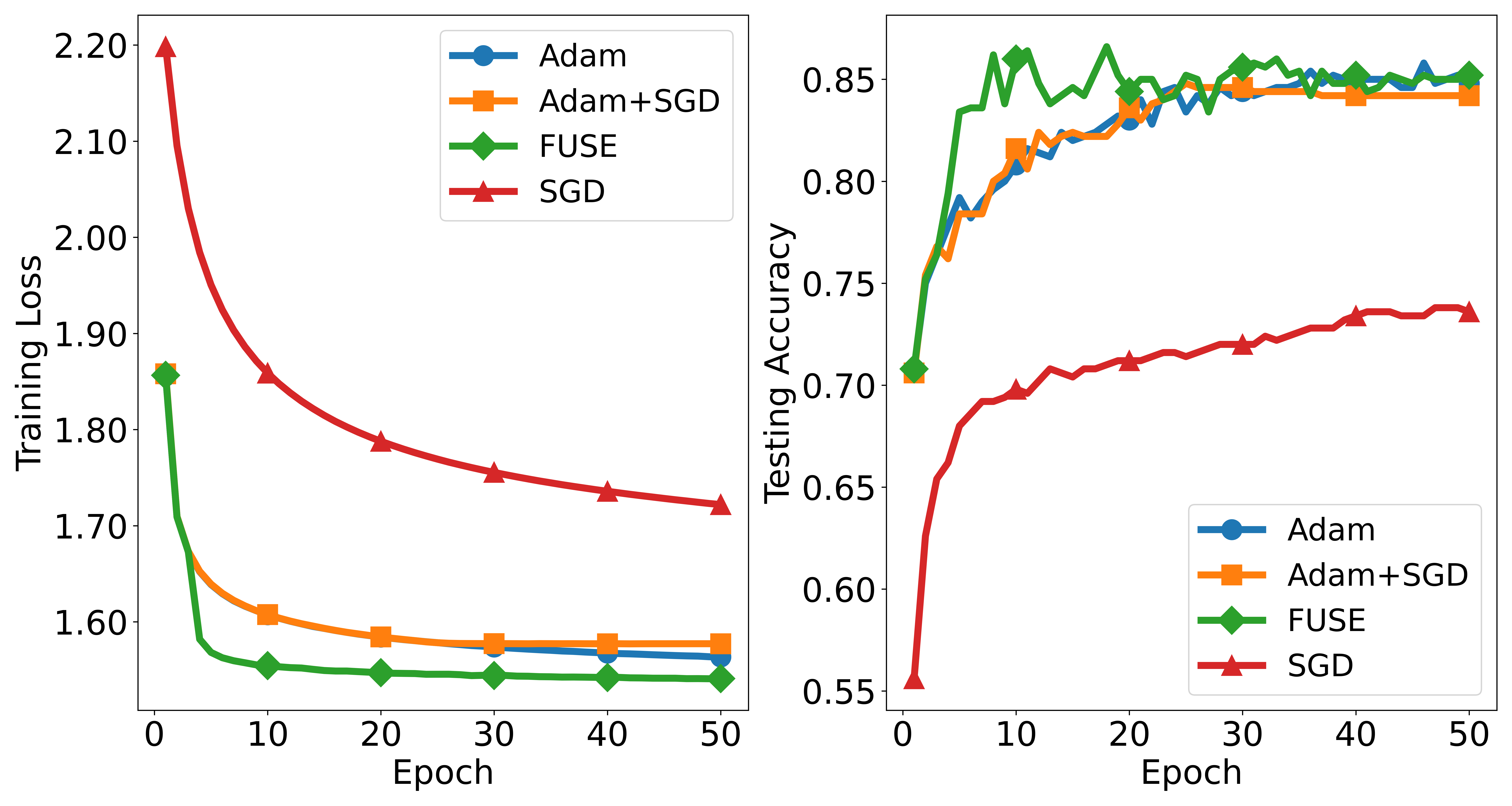}
    \caption{Logistic Regression performance on FashionMNIST}
    \label{fig:LR_FashinMNIST}
\end{figure}

\begin{figure}[ht]
    \centering
    \captionsetup{skip=-1pt}
    \includegraphics[height=4cm,width=7cm]{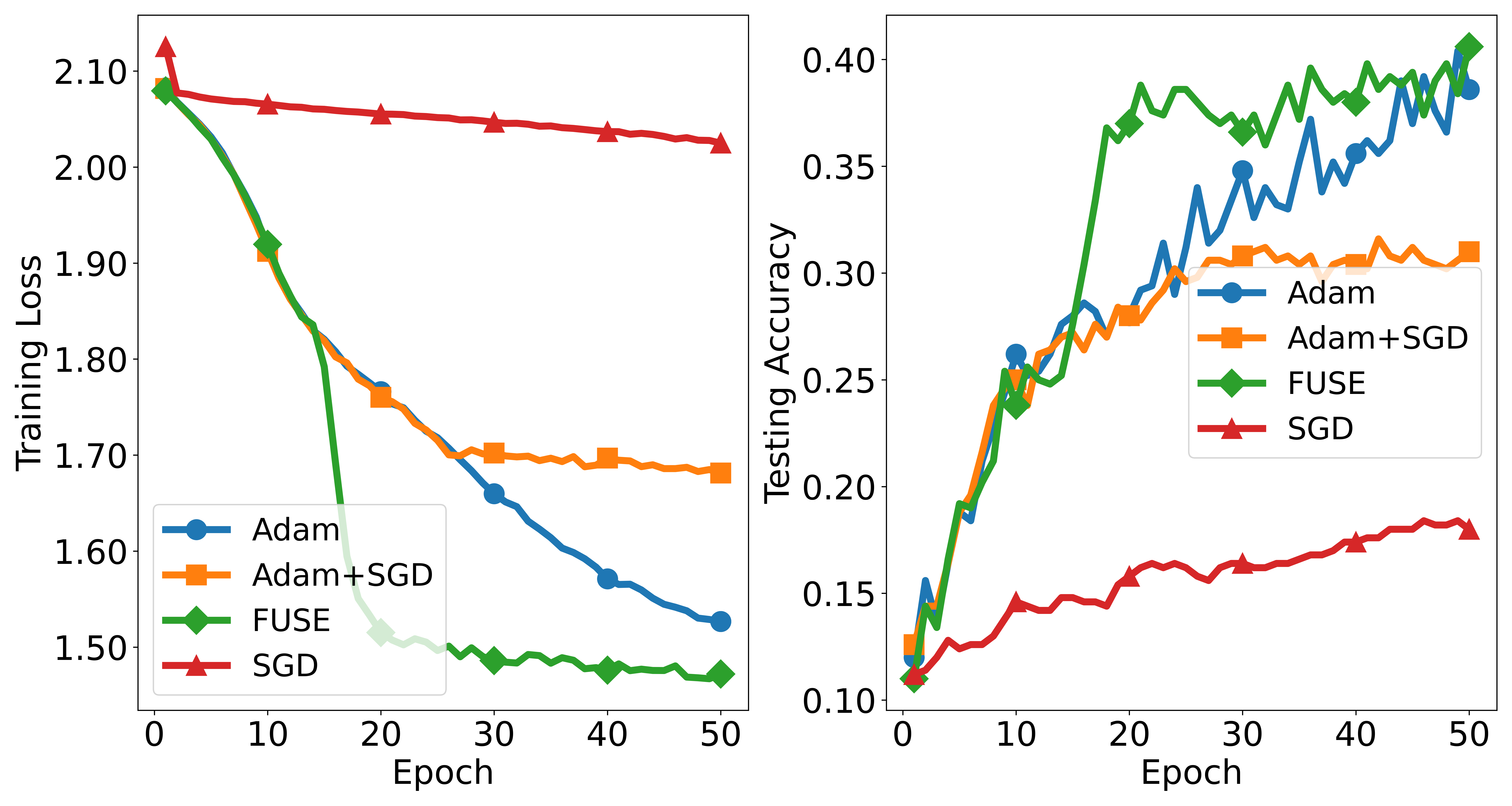}
    \caption{DenseNet performance on CIFAR-10}
    \label{fig:cifar_Dnet}
\end{figure}

\noindent\textbf{Real datasets with diverse models.}
Table~\ref{table:model_results} shows testing performance for a variety of models and datasets. 
For a few cases, the standard deviation is close to 0 ($\sim0$) with different random seeds. Overall, \texttt{FUSE} outperforms all baselines. In Fig.~\ref{fig:criterion}, different switchover criteria are used in the \texttt{FUSE} optimizer and the result indicates that the epoch-based criterion slightly outperforms the other two. Hence, we use this simple criterion, while determining the nearly optimal epoch number by manual tuning.
However, it can be regarded as a hyperparameter to be tuned as others in regular deep learning models.
Fig.~\ref{fig:LR_FashinMNIST} and Fig.~\ref{fig:cifar_Dnet} show the training loss and testing accuracy w.r.t. the number of epochs. Evidently, \texttt{FUSE} leads to improvement over baselines since the initialization by \texttt{Adam} helps generalization in testing data. However, we also notice that \texttt{FUSE} may result in slightly higher variance relative to first-order methods, particularly after switching to the second-order method. This is primarily attributed to the stochasticity in approximating the Hessian, which can be mitigated by increasing the batch size. Another interesting observation from both figures is that though given a sufficient number of epochs, \texttt{Adam} can still be on par with \texttt{FUSE}. But given the limited computational budget, \texttt{FUSE} ensures a faster convergence speed and enables computational cost reduction.
In summary, our proposed \texttt{FUSE} is able to outperform the popular first-order optimizers, particularly when computing resources are insufficient. Additionally, training different model architectures also necessitates a suitable optimizer selection instead of sticking to only first-order optimizers consistently.



\section{Conclusion}
In this paper, we present a novel algorithm that synthesizes both first-order and second-order algorithms in a unified framework. Specifically, we develop \texttt{FUSE} and \texttt{FUSE-PV} based on \texttt{Adam} and \texttt{L-BFGS} optimization algorithms. We also introduce different metrics for determining when to switch from first-order to second-order algorithms. We theoretically analyze the computational complexity for \texttt{FUSE-PV}. To validate our proposed method, simple non-convex functions and multiple models with diverse datasets are evaluated, and the results show superiority over diverse baselines.
\vspace{3pt}



\bibliographystyle{IEEEtran}
\bibliography{main}

\begin{thebibliography}{10}
\providecommand{\url}[1]{#1}
\csname url@samestyle\endcsname
\providecommand{\newblock}{\relax}
\providecommand{\bibinfo}[2]{#2}
\providecommand{\BIBentrySTDinterwordspacing}{\spaceskip=0pt\relax}
\providecommand{\BIBentryALTinterwordstretchfactor}{4}
\providecommand{\BIBentryALTinterwordspacing}{\spaceskip=\fontdimen2\font plus
\BIBentryALTinterwordstretchfactor\fontdimen3\font minus \fontdimen4\font\relax}
\providecommand{\BIBforeignlanguage}[2]{{%
\expandafter\ifx\csname l@#1\endcsname\relax
\typeout{** WARNING: IEEEtran.bst: No hyphenation pattern has been}%
\typeout{** loaded for the language `#1'. Using the pattern for}%
\typeout{** the default language instead.}%
\else
\language=\csname l@#1\endcsname
\fi
#2}}
\providecommand{\BIBdecl}{\relax}
\BIBdecl

\bibitem{bottou2012stochastic}
L.~Bottou, ``Stochastic gradient descent tricks,'' in \emph{Neural networks: Tricks of the trade}.\hskip 1em plus 0.5em minus 0.4em\relax Springer, 2012, pp. 421--436.

\bibitem{defossez2020convergence}
A.~D{\'e}fossez, L.~Bottou, F.~Bach, and N.~Usunier, ``On the convergence of adam and adagrad,'' \emph{arXiv e-prints}, pp. arXiv--2003, 2020.

\bibitem{zou2019sufficient}
F.~Zou, L.~Shen, Z.~Jie, W.~Zhang, and W.~Liu, ``A sufficient condition for convergences of adam and rmsprop,'' in \emph{Proceedings of the IEEE/CVF Conference on Computer Vision and Pattern Recognition}, 2019, pp. 11\,127--11\,135.

\bibitem{savarese2019convergence}
P.~Savarese, ``On the convergence of adabound and its connection to sgd,'' \emph{arXiv preprint arXiv:1908.04457}, 2019.

\bibitem{liu2019variance}
L.~Liu, H.~Jiang, P.~He, W.~Chen, X.~Liu, J.~Gao, and J.~Han, ``On the variance of the adaptive learning rate and beyond,'' \emph{arXiv preprint arXiv:1908.03265}, 2019.

\bibitem{boyd2004convex}
S.~P. Boyd and L.~Vandenberghe, \emph{Convex optimization}.\hskip 1em plus 0.5em minus 0.4em\relax Cambridge university press, 2004.

\bibitem{zhu1997algorithm}
C.~Zhu, R.~H. Byrd, P.~Lu, and J.~Nocedal, ``Algorithm 778: L-bfgs-b: Fortran subroutines for large-scale bound-constrained optimization,'' \emph{ACM Transactions on mathematical software (TOMS)}, vol.~23, no.~4, pp. 550--560, 1997.

\bibitem{polyak1969conjugate}
B.~T. Polyak, ``The conjugate gradient method in extremal problems,'' \emph{USSR Computational Mathematics and Mathematical Physics}, vol.~9, no.~4, pp. 94--112, 1969.

\bibitem{xu2020second}
P.~Xu, F.~Roosta, and M.~W. Mahoney, ``Second-order optimization for non-convex machine learning: An empirical study,'' in \emph{Proceedings of the 2020 SIAM International Conference on Data Mining}.\hskip 1em plus 0.5em minus 0.4em\relax SIAM, 2020, pp. 199--207.

\bibitem{yao2020adahessian}
Z.~Yao, A.~Gholami, S.~Shen, M.~Mustafa, K.~Keutzer, and M.~W. Mahoney, ``Adahessian: An adaptive second order optimizer for machine learning,'' \emph{arXiv preprint arXiv:2006.00719}, 2020.

\bibitem{anil2020scalable}
R.~Anil, V.~Gupta, T.~Koren, K.~Regan, and Y.~Singer, ``Scalable second order optimization for deep learning,'' \emph{arXiv preprint arXiv:2002.09018}, 2020.

\bibitem{anil2020second}
------, ``Second order optimization made practical,'' \emph{arXiv preprint arXiv:2002.09018}, 2020.

\bibitem{ma2020apollo}
X.~Ma, ``Apollo: An adaptive parameter-wise diagonal quasi-newton method for nonconvex stochastic optimization,'' \emph{arXiv preprint arXiv:2009.13586}, 2020.

\bibitem{jin2021exploiting}
Q.~Jin and A.~Mokhtari, ``Exploiting local convergence of quasi-newton methods globally: Adaptive sample size approach,'' \emph{arXiv preprint arXiv:2106.05445}, 2021.

\bibitem{di2023lsos}
D.~Di~Serafino, N.~Kreji{\'c}, N.~Krklec~Jerinki{\'c}, and M.~Viola, ``Lsos: Line-search second-order stochastic optimization methods for nonconvex finite sums,'' \emph{Mathematics of Computation}, vol.~92, no. 341, pp. 1273--1299, 2023.

\bibitem{arjevani2020second}
Y.~Arjevani, Y.~Carmon, J.~C. Duchi, D.~J. Foster, A.~Sekhari, and K.~Sridharan, ``Second-order information in non-convex stochastic optimization: Power and limitations,'' in \emph{Conference on Learning Theory}.\hskip 1em plus 0.5em minus 0.4em\relax PMLR, 2020, pp. 242--299.

\bibitem{niu2023ml}
Y.~Niu, Z.~Fabian, S.~Lee, M.~Soltanolkotabi, and S.~Avestimehr, ``ml-bfgs: A momentum-based l-bfgs for distributed large-scale neural network optimization,'' \emph{arXiv preprint arXiv:2307.13744}, 2023.

\bibitem{keskar2020improving}
N.~S. Keskar and R.~Socher, ``Improving generalization performance by switching from adam to sgd. 2017,'' \emph{arXiv preprint arXiv:1712.07628}, 2020.

\bibitem{landro2020mixing}
N.~Landro, I.~Gallo, and R.~La~Grassa, ``Mixing adam and sgd: a combined optimization method,'' \emph{arXiv preprint arXiv:2011.08042}, 2020.

\bibitem{wang2020adasgd}
J.~Wang and J.~Wiens, ``Adasgd: Bridging the gap between sgd and adam,'' \emph{arXiv preprint arXiv:2006.16541}, 2020.

\bibitem{shi2020rmsprop}
N.~Shi, D.~Li, M.~Hong, and R.~Sun, ``Rmsprop converges with proper hyper-parameter,'' in \emph{International Conference on Learning Representations}, 2020.

\bibitem{bottou2018optimization}
L.~Bottou, F.~E. Curtis, and J.~Nocedal, ``Optimization methods for large-scale machine learning,'' \emph{SIAM review}, vol.~60, no.~2, pp. 223--311, 2018.

\bibitem{berahas2016multi}
A.~S. Berahas, J.~Nocedal, and M.~Tak{\'a}c, ``A multi-batch l-bfgs method for machine learning,'' \emph{Advances in Neural Information Processing Systems}, vol.~29, 2016.

\bibitem{vaswani2019fast}
S.~Vaswani, F.~Bach, and M.~Schmidt, ``Fast and faster convergence of sgd for over-parameterized models and an accelerated perceptron,'' in \emph{The 22nd international conference on artificial intelligence and statistics}.\hskip 1em plus 0.5em minus 0.4em\relax PMLR, 2019, pp. 1195--1204.

\bibitem{wang2019sadam}
G.~Wang, S.~Lu, W.~Tu, and L.~Zhang, ``Sadam: A variant of adam for strongly convex functions,'' \emph{arXiv preprint arXiv:1905.02957}, 2019.

\bibitem{xiao2017fashion}
H.~Xiao, K.~Rasul, and R.~Vollgraf, ``Fashion-mnist: a novel image dataset for benchmarking machine learning algorithms,'' \emph{arXiv preprint arXiv:1708.07747}, 2017.

\bibitem{kasun2016dimension}
L.~L.~C. Kasun, Y.~Yang, G.-B. Huang, and Z.~Zhang, ``Dimension reduction with extreme learning machine,'' \emph{IEEE transactions on Image Processing}, vol.~25, no.~8, pp. 3906--3918, 2016.

\bibitem{shehab2017survey}
M.~Shehab, A.~T. Khader, and M.~A. Al-Betar, ``A survey on applications and variants of the cuckoo search algorithm,'' \emph{Applied soft computing}, vol.~61, pp. 1041--1059, 2017.

\end{thebibliography}

\end{document}